%% file: root.tex
\documentclass[letterpaper, 10 pt, conference]{ieeeconf}  

\IEEEoverridecommandlockouts                              

\usepackage{graphicx}
\usepackage{algorithm}
\usepackage{colortbl}
\usepackage{xcolor}
\usepackage{booktabs}
\usepackage{amsmath}
\usepackage{amssymb}
\usepackage{booktabs}
\usepackage{paralist}
\usepackage{epsfig}
\usepackage{wrapfig,lipsum}
\usepackage{tabularx}
\usepackage{makecell}
\usepackage{hyperref}
\usepackage{multirow}

\input{text/macro}

\title{\LARGE \bf
Prompting Multi-Modal Tokens to Enhance End-to-End Autonomous Driving Imitation Learning with LLMs }

\author{Yiqun Duan$^{1}$$^{,}$$^{*}$, Qiang Zhang$^{2}$$^{,}$$^{*}$, Renjing Xu$^{2}$$^{,}$$^{\dagger}$ 
\thanks{$^{1}$Yiqun Duan is with the HAI Centre, Australia Artificial Intelligence Institute, School of Computer Science, University of Technology Sydney, 2007, Ultimo, Australia
        {\tt\small yiqun.duan@student.uts.edu.au}}%
\thanks{$^{2}$ Qiang Zhang and Renjing Xu are with the School of Computer Science at The Hong Kong University of Science and Technology (Guangzhou)
        {\tt\small qzhang749@connect.hkust-gz.edu.cn, renjingxu@hkust-gz.edu.cn}}
\thanks{
        $^{*}$ are equal contributors, ${\dagger}$ is the corresponding author. This paper is published as an oral presentation paper at the 2024 IEEE International Conference on Robotics and Automation (ICRA2024), Yokohama, Japan.}%
}

\begin{document}

\maketitle
\thispagestyle{empty}
\pagestyle{empty}

\begin{abstract}

The utilization of Large Language Models (LLMs) within the realm of reinforcement learning, particularly as planners, has garnered a significant degree of attention in recent scholarly literature. However, a substantial proportion of existing research predominantly focuses on planning models for robotics that transmute the outputs derived from perception models into linguistic forms, thus adopting a `pure-language' strategy. 
In this research, we propose a hybrid End-to-End learning framework for autonomous driving by combining basic driving imitation learning with LLMs based on multi-modality prompt tokens. 
Instead of simply converting perception results from the separated train model into pure language input, our novelty lies in two aspects. 
1) The end-to-end integration of visual and LiDAR sensory input into learnable multi-modality tokens, thereby intrinsically alleviating description bias by separated pre-trained perception models. 
2) Instead of directly letting LLMs drive, this paper explores a hybrid setting of letting LLMs help the driving model correct mistakes and complicated scenarios.
The results of our experiments suggest that the proposed methodology can attain driving scores of $49.21\%$, coupled with an impressive route completion rate of $91.34\%$ in the offline evaluation conducted via CARLA. These performance metrics are comparable to the most advanced driving models. 

\end{abstract}


\input{text/intro}

\input{text/relatedworks}

\input{text/methodology}

\input{text/experiment}

\section*{ACKNOWLEDGMENT}

We would like to extend our heartfelt thanks to the Beijing Innovation Center of Humanoid Robotics Co., Ltd. for their substantial support in our research endeavors.

\bibliographystyle{ieeetr}
\bibliography{root}  







\end{document}

%% file: text/macro.tex
\newcommand{\bA}{\mathbf{A}}

\newcommand{\bF}{\mathbf{F}} %

\newcommand{\bK}{\mathbf{K}}

\newcommand{\bQ}{\mathbf{Q}}

\newcommand{\bV}{\mathbf{V}}
\newcommand{\bw}{\mathbf{w}}
\newcommand{\bx}{\mathbf{x}}

\newcommand{\cL}{\mathcal{L}}

\makeatletter
\DeclareRobustCommand\onedot{\futurelet\@let@token\@onedot}
\def\@onedot{\ifx\@let@token.\else.\null\fi\xspace}

\makeatother



%% file: text/intro.tex
\section{Introduction}

Recent advancements in AI systems for autonomous driving can be classified into two main categories: Pipeline Formation and End-to-End (E2E) Formation.
Pipeline formation~\cite{gog2021pylot,liu2020hercules,jiao2021greedy,liu2021ground,li2020lidar, song2020pip,wen2020scenario,claussmann2019review}, involves manually decomposing the drive into sequential modules. These modules consist of tasks such as localization \cite{qin2021light,woo2018localization}, scene reconstruction \cite{dyer2001volumetric,bozic2021transformerfusion}, planning \cite{letchner2006trip,abu2022comprehensive}, and control \cite{tsugawa1994vision,kong2015kinematic} according to certain rules. 
However, these separately learned rules struggle to cover long-tail scenarios, which are crucial in driving situations. Thus, researchers are aiming to make the system learn using an end-to-end formation through reinforcement or imitation learning, to better mimic human-like decision-making and bring it a step closer to handling these important scenarios effectively.
\begin{figure*}[t]
    \centering
    \includegraphics[width=0.8\textwidth]{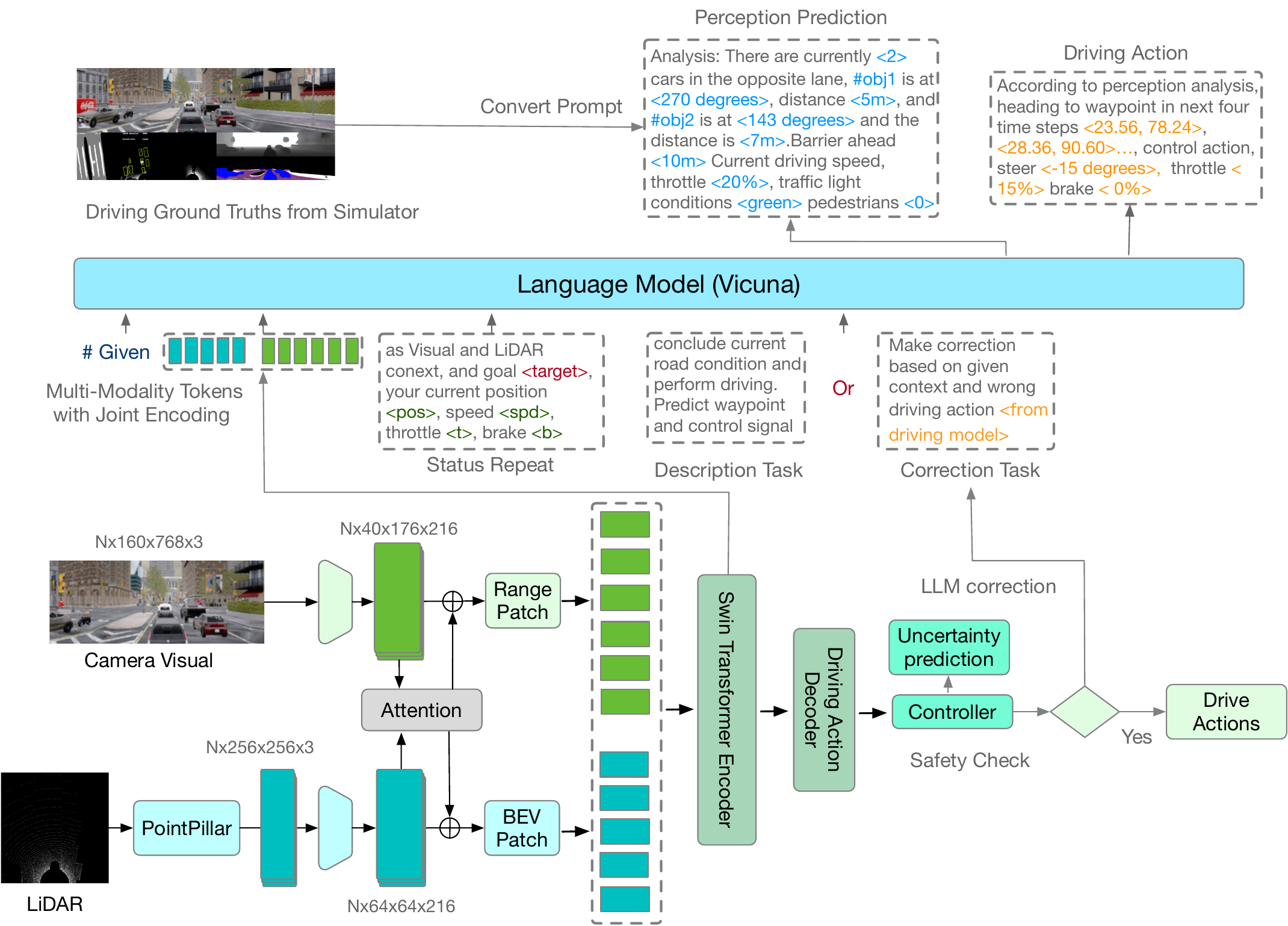}
    \vspace{-5pt}
    \caption{Overview of utilizing language model for autonomous driving. The system takes the camera and LiDAR input and extracts shallow feature maps with two branches. Then a joint Swin transformer encodes perception input into joint token representation. The prompt is constructed by sequentially concatenating multi-modality joint tokens, status repeat tokens, and driving task tokens. The task prompts consist of two kinds 1) directly predicting the driving actions and 2) driving action correction given the driving output. The controller model also predicts an uncertainty score by an MLP layer, which decides whether to ask GPT to correction or drive by itself. 
    The model is trained by auto-regressively predicting perception description and the driving action. Driving actions are executed by a final controller.}
    \label{fig:overview}
    \vspace{-5pt}
\end{figure*}

E2E driving approaches~\cite{hu2022st,tampuu2020survey,NVIDIA-2016,chitta2021neat,toromanoff2020end} employ state-to-action imitation or reinforcement learning strategies on intermediate feature representation states. This enables agents to behave appropriately in various driving contexts.
The introduction of multi-modal sensory fusion techniques \cite{transfuser,zhang2022mmfn,huang2021bevdet,liu2022bevfusion} have significantly improved the perception state representation. These techniques make use of both visual and LiDAR data \cite{fadadu2022multi,chen2017multi,zhou2020end,huang2021bevdet,liu2022bevfusion,li2022bevdepth}, presenting it in a Bird's Eye View (BEV) format or as a flat tensor for the decoding process. ST-P3~\cite{hu2022st} approach and Transfuser~\cite{transfuser} reach SOTA performance with the multi-modality imitation learning. 

However, these methods emphasize on state representation and pay less attention to the decoder. Generally, the learning system predicts waypoints for the next n time steps based on the encoded feature, with an additional controller translating waypoints into actual controls.
However, merely predicting waypoints falls short in the face of the complex demands of autonomous driving. This limitation hinders the practical application of end-to-end driving systems. 
Interfuser~\cite{shao2022safety} and TCP \cite{wu2022trajectoryguided} respectively improve the decoder by introducing safety constraints on the controller and directly adding a branch to predict safety control signals. 
UniAD~\cite{hu2023_uniad} try to unify pipelines with an end-to-end training framework and reach SOTA performance by introducing more immediate supervision such as occupancy and lanes. 
Especially, ThinkTwice~\cite{jia2023thinktwice} proposes to involve the decoder in an anthropomorphized approach that enables interaction between the encoder and decoder following `driving after thinking twice'.

However, it's hard to manually list all possible driving rules or components comprehensively. In that case, letting the autonomous system learn semantically could benefit the end-to-end driving tasks.  
ADAPT~\cite{jin2023adapt} has given a first exploration of modeling the driving task as an image captioning model, which simultaneously predicts driving actions and driving explanation as a caption using visual transformers. 
More recent works have investigated using Large Language Models (LLMs) for wider autonomous/robotic systems. 
PaLM-E~\cite{driess2023palme} utilized pre-trained LLM to complete multiple embodied tasks, including sequential robotic manipulation planning, visual question answering, and captioning. 
Mini-GPT4~\cite{zhu2023minigpt} and LLaVA~\cite{liu2023llava} introduce practical approaches for visual context tuning for LLMs.
However, end-to-end autonomous driving has not yet been adequately explored. Moreover, these methods concentrate on the language model in an auto-regressive way. How to incorporate imitation learning and reinforcement learning in autonomous driving scenarios remain uninvestigated. 

In this paper, we propose to model end-to-end autonomous driving as a multi-modality language model. 
For state encoding, we propose a two-branch framework to encode both visual and LiDAR features into joint feature token representations. 
We prompt the multi-modality visual tokens into driving languages, which contain driving descriptions followed by driving actions, not only waypoints but also control signals. 
The driving-like language formation could help the model learn comprehensive driving skills not merely waypoints, not merely control signals, but also the reasoning before taking the action. 
Despite the auto-regression loss, we design a subtle mechanism to incorporate a language model with driving metrics reward-guided reinforcement tuning. 
Considering the randomness of the large language generation model, 1) the generated results may exceed the normal safety threshold or do not contain waypoints or driving actions. 
2) the generated waypoints and the control signal may conflict with each other.
To enhance driving safety, we design a mechanism of re-quering the driving decision with enhanced prompt to enable model to rethink the driving decision. 

We have performed comprehensive experiments using the off-board driving simulator CARLA~\cite{carla}, a platform capable of providing the requisite driving information necessary for generating prompted language supervision. The findings of our experiments suggest that LLMs can attain a level of driving performance comparable to the current state-of-the-art models in this domain.
The incorporation of language models into driving systems with reinforcement learning enhancement offers clear advantages. When driving actions are prompted through language, it enables model learning that extends beyond the mere prediction of waypoints. It also allows for the extraction of the underlying driving logic that is inherently contained within the language itself. This methodology serves as a step towards the development of a more personalized driving model, thereby bringing us closer to the goal of a more human-like autonomous driving system.
The contribution lies threefold.
\begin{compactitem}
    \item A novel framework incorporates multi-modality perception as joint token representations into language prompts for end-to-end autonomous driving. 
    \item A simple yet effective prompting approach on unified observation, current states, trajectory, and control actions in continuous prompts. 
    \item Incorporating reward-guided reinforcement learning supervision on language prompts in autonomous driving scenarios.
\end{compactitem}

%% file: text/relatedworks.tex
\section{Related Works}
\textbf{Imitation Learning for End-to-End Driving:} The evolution of End-to-End autonomous driving has typically followed a \textit{destructured, then structured} trajectory. Early explorations, such as ALVINN~\cite{pomerleau1988alvinn}, and DAVE~\cite{bojarski2016end}, modeled a simple projection relationship between the input from a single-view camera and the steering wheel or accelerator angle. The prevalent approach at this stage was to eliminate human prior knowledge, thereby offering the network a more direct learning target to improve the fitting capability for fully supervised behavior cloning.
However, sensory input was limited in this early era, leading to the introduction of multiple imitation learning techniques~\cite{codevilla2018end, bansal2018chauffeurnet,kendall2019learning} to enhance the data quality for behavior cloning. Subsequent developments saw the expansion of camera sensor usage \cite{hecker2018end} from a single view to multiple views to enhance planning stability.
In more recent works, the reintroduction of prior knowledge has been explored to augment driving performance. WOR \cite{wor} and MARL\cite{marl} suggest that incorporating prior trained knowledge from related vision tasks such as detection~\cite{resnet,CNN,yolov3} could enhance model performance. NEAT~\cite{chitta2021neat} introduced an attention module between different camera views to boost feature quality. ST-P3~\cite{hu2022st}, VAD~\cite{jiang2023vad}, and UniAD~\cite{hu2023_uniad} amalgamate human prior module design from pipeline methods to increase driving rationality.

\textbf{Reinforcement Learning for LLMs:}\quad 
For language models, relying solely on supervised learning to generate accurate responses is inadequate as it's difficult to determine whether the model truly understands the answer to a question. The RL framework motivates a model to produce correct answers in a natural manner without resorting to fabrication because the model receives poor scores for incorrect answers. Reinforcement Learning from Human Feedback (RLHF) \cite{christiano2017deep} is a critical technology in the current wave of large language models (LLMs). Three of OpenAI's most prominent LLMs, namely InstructGPT \cite{ouyang2022training}, ChatGPT, and GPT4, rely on RLHF as their core technology. Although PaLM-E integrated the 540 billion-parameter PaLM \cite{chowdhery2022palm} LLM and the 22 billion-parameter Vision Transformer \cite{dehghani2023scaling} to create an embodied multimodal language model, its approach on the autonomous driving task lacked sufficient detail. Haomo's DriveGPT~\footnote{\href{https://drivegpt.haomoai.com/}{https://drivegpt.haomoai.com/}} for the autonomous driving scenario proposes to convert BEV information into drive language and learning language tokens.
However, it is not solely end-to-end as it does not learn directly with raw tokens from sensors such as cameras and LiDAR.

\textbf{Multi-Modality Fusion For Driving Tasks:} The limited capabilities of a monotonous camera sensor have led to the exploration of other sensory modalities, such as LiDAR, Radar, and Openmap, to enhance driving stability~\cite{pointpillars,rhinehart2019precog,vectornet,lbc}.
Most previous approaches to modality fusion have been specifically designed for certain perception tasks such as 3D object detection~\cite{huang2021bevdet,liu2022bevfusion}, depth estimation~\cite{li2022bevdepth,zhao2020monocular}, and instance motion forecasting~\cite{hu2021fiery,luo2018fast}. BEVFusion~\cite{liu2022bevfusion}, for instance, fuses LiDAR with image features by using completely independent encoders with geometric projection into the same Bird's-eye-view (BEV) feature space at a late stage.
However, Transfuser~\cite{transfuser} observed that pure geometrical fusion representation impairs the performance of comprehensive and complex urban E2E autonomous driving.

\vspace{-1pt}

%% file: text/methodology.tex
\section{Methodology}

\subsection{Multi-Modality Joint Token Encoder}
In order to provide a more suitable driving context for GPT-based models, we propose to fuse Visual-LiDAR perception input into a joint token representation. 
The architecture is described in Fig.~\ref{fig:overview}, which contains two stages, \textit{early fusion} and \textit{late fusion}. This two-stage network is designed as follows:

\textbf{Early Fusion:} In the first stage, two distinct CNN branches are utilized to extract shallow features from a monotonic image and LiDAR inputs, respectively. The image branch concatenates three front view camera inputs, each with a field of view of 60 degrees, into a single monotonic view, subsequently reshaped to the shape $3\times160\times704$. On the other hand, the LiDAR branch processes raw LiDAR input with PointPillar~\cite{pointpillars} to create a bird's eye view (BEV) feature with shape $33\times256\times256$. As these lower-level features retain strong geometric relations, the independent encoder can extract tight local feature representation with fewer distractions. During this stage, we apply cross-modality self-attention to enhance geometrical feature fusion. 

The feature map $\bF_{im}$, $\bF_{Li}$ representing the image branch and LiDAR branch are reshaped by flattening~\cite{dosovitskiy2020image,transfuser} and concatenating into sequence $\bF_{in} = \rm{cat}(\bF_{im}, \bF_{Li})$. 
The transformer layer is applied on $\bF_{in} \in \mathbb{R}^{N^\star \times C}$ to perform multi-head attention between each token in $N^\star$ dimension as defined in Eq.~\ref{eq:self_out}.

\begin{equation}\label{eq:self_out}
    \begin{aligned}
    \bQ_F = &~ \bF_{in} W_Q, \ \bK_F = \bF_{in} W_K, \ \bV_F = \bF_{in} W_V,\\
    \bF_{Li} = &~ \rm{m_1}(\bA)+\bF_{in}, \ \bF_{im}=\rm{m_2}(\bA)+\bF_{in}, \\
  \bA = &~ \rm{softmax}(\frac{\bQ_F \bK_F^T}{\sqrt{D}})\bV_F
    \end{aligned} 
\end{equation}
where $\rm{m_1}$ and $\rm{m_2}$ denote two different MLP layers that project the joint attention into proper shape and are directly added to the original feature map after normalization.

\textbf{Late Fusion:} 
After low-level feature extraction with attention, we intend to align features from both modalities into a unified semantic token space, where we treat each $16\times 16 \times C$ feature segment to be a semantic \textit{word} as shown in Fig.~\ref{fig:overview}. 
To maintain the spatial relationships across these tokens, we use the cross-attention mechanism.
\begin{figure}[htbp]
  \centering
  \begin{minipage}[t]{1.0\columnwidth}
    \includegraphics[width=\columnwidth]{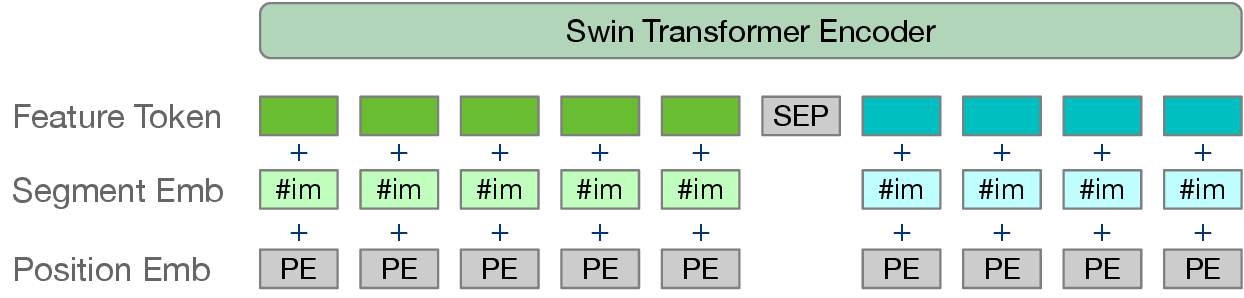}
    \label{fig:token}
  \end{minipage}

    \caption{Illustration of joint token representation, the perception token is aligned with segmentation embedding and position embedding to distinguish from normal word token.  }
\end{figure}

A segment embedding is added with positional embedding $\mathbf{PE}+\mathbf{SE}$ to indicate the position information is from range view or BEV feature space.
The feature maps from different modalities are tokenized and concatenated into unified tokens $\bF_{s} \in \mathbb{R}^{N^\star \times C}$ with patch size $16\times16$ by adding $\mathbf{PE}+\mathbf{SE}$ to the token. 
A 4-layer transformer encoder is applied on the unified token sequence $\bF_{s}$ by setting query $Q(\bF_{s})$, key $K(\bF_{s})$, and value $V(\bF_{s})$. 
As the late fusion is performed by share transformer encoding, this could introduce a better joint semantic feature representation between especially for the language model. 

\subsection{Prompt Construction}
\label{subsec:prompt}
In the endeavor to integrate Large Language Models (LLMs) within the domain of autonomous driving, a salient challenge lies in the effective design of language prompts and the translation of supervisory signals into language constructs. A key factor in this process entails the creation of prompts that adeptly incorporate multimodal tokens, while also considering the vehicle's instantaneous state, the ambient environment, and short-term as well as long-term objectives.
Instead of letting the LLMs directly drive, the basic design logic is to let LLMs help the basic driving model correct driving actions.
This setting alleviates the LLMs generating actions or goals not exist in real scenarios.

\begin{figure}[hbpt]
    \centering
    \includegraphics[width=1.0\columnwidth]{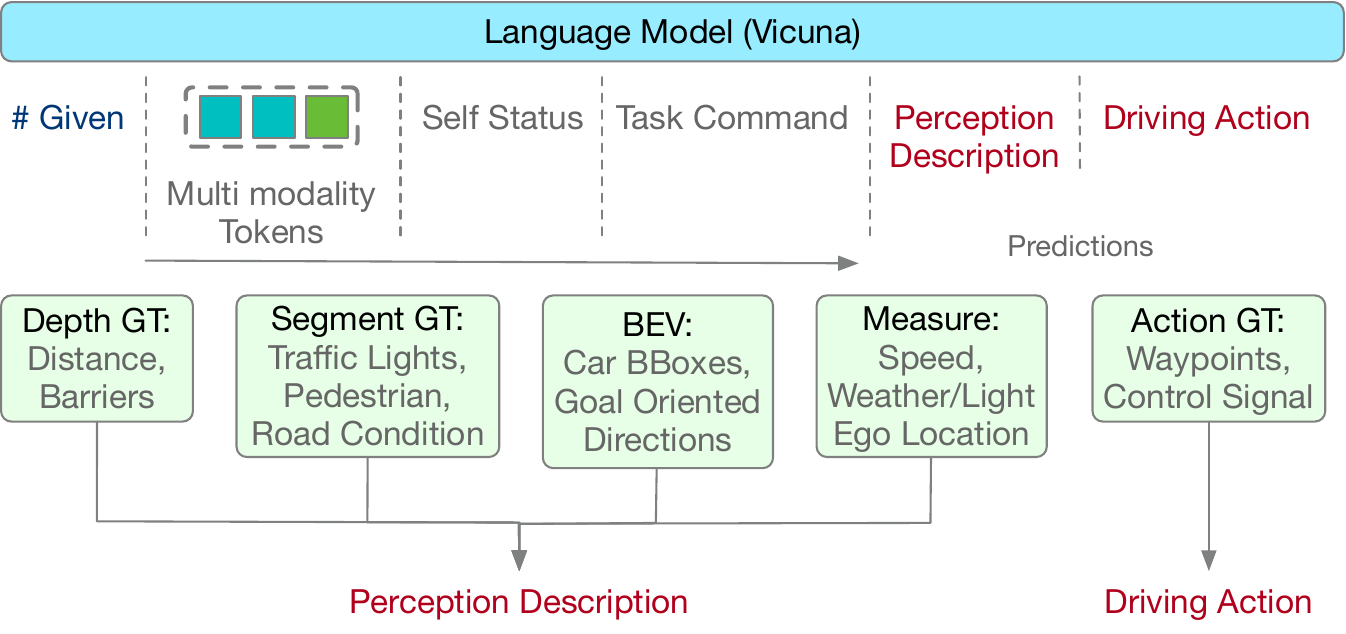}
    \caption{Illustration of the prompt construction. Both traditional perception supervision and driving actions are converted into language descriptions through a descriptor.  }
    \label{fig:prompt}
\end{figure}

The given information during driving is first contacted sequentially, 1) multi-modality tokens 2) self-status of the car 3) driving task command. 
Here, the self-status of the car denotes the current speed, throttle, brake, and current position and driving task commands denote the instruction (perform driving/solve drive conflicts).
After that, supervisions are appended sequentially as language guidance considering two aspects 1) perception description and 2) driving action as shown in Fig.~\ref{fig:prompt}.
Perception description guides the language model to predict perception results which are mostly used as auxiliary heads supervision in previous works~\cite{transfuser,nie2020mmfn,jiang2023vad, hu2023_uniad}, such as segmentation, depth prediction, BEV prediction, and detection. 
Driving action predicts both the waypoints in the next 4 time steps and control actions. 
A notable advantage of auto-regressive models over their traditional counterparts is their inherent ability to learn driving logic, facilitated by their intrinsic capacity to assimilate the complex, underlying linguistic logic. 
Towards this end, we have proposed a comprehensive prompt design strategy, which is depicted above. 

\textbf{Tasks:} The task prompts are designed to have three modes. 1) Given the multi-modality tokens directly generate perception observation output, and driving actions. Normally the driving language output is not used for driving considering the driving cost, yet the normal learning parallel with the driving model is crucial for establishing driving logic. 
2) Re-query the LLM if the driving modal's language output contradicts to the safety controller. 
3) Given the multi-modality tokens and driving output, correct the driving error by LLMs.

\subsection{Re-Query Mechanism} 

\begin{figure}
    \centering
    \includegraphics[width=1.0\columnwidth]{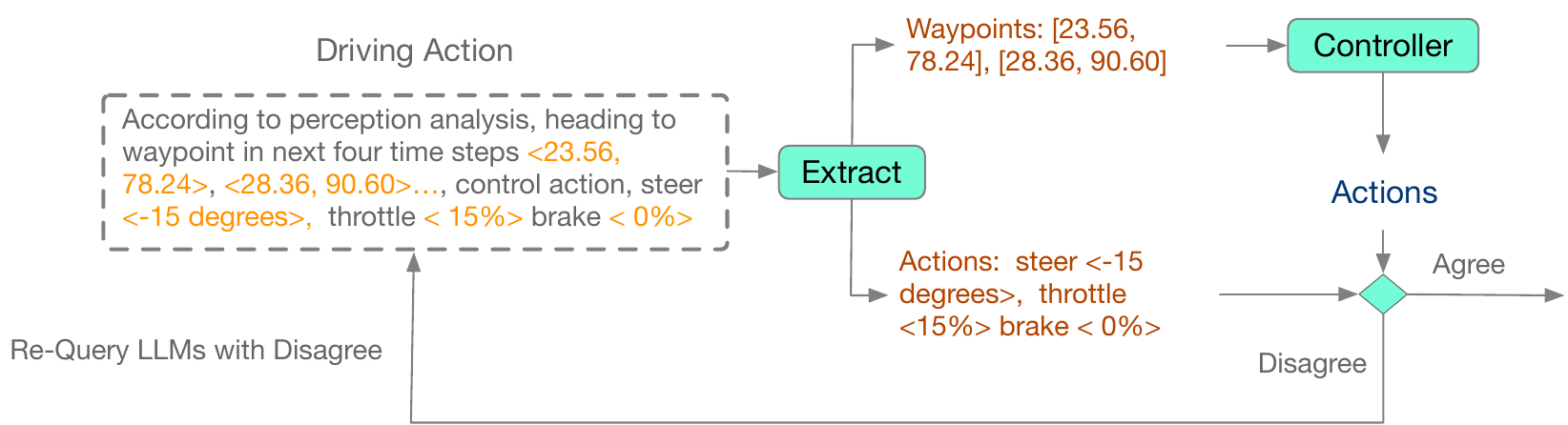}
    \caption{Illustration of the re-query mechanism when there is disagreement.\label{fig:re-query}}
    \vspace{-10pt}
\end{figure}

A remarkable problem of auto-regressive models hiders the performance is the uncertainty of the language prediction.
Since in Section~\ref{subsec:prompt}, we propose to predict comprehensive driving actions and waypoints together in sequential order, it is observed in our experiment that the waypoints prediction might be conflicted with the control actions as illustrated in Fig.~\ref{fig:re-query}. 
Here the driving actions and waypoints are extracted through predefined regular expressions. 
The waypoints are converted to the control signal by the traditional pid controller used in previous works~\cite{transfuser,nie2020mmfn}. 
If conflicts between the predicted action and the exceed the pre-defined threshold the system will trigger a re-query mechanism. 
At this stage, the task commands part in the prompt (Fig.~\ref{fig:overview}) will be changed from the primary driving commands to query with the previous disagreement. 
If without an agreement, we will select the waypoints leaded action for final driving control. 
This mechanism could help the language model to `think twice' when dealing with unclear or hard situations.

\subsection{Reinforcement Guided Tuning}

Pure auto-regressive learning is not sufficient to deal with the regression trajectories. Yet, under our design, the waypoints are purely learned in auto-regressive with MLE loss. In that case, we add a reinforcement-guided regression loss to boost the model with the prediction accuracy. 
Different from InstructGPT~\cite{ouyang2022training} that need human feedback to apply reinforcement fine-tuning, our scenarios have automated generated guidance. 
As the goal is to imitate the expert driving trajectories which are accessible in our task setting, we can assume the description of the expert trajectories as $y_w$. 
Since the GPT-based model only could put supervision in token wise yet, while the whole sequence is meaningful for regression, we introduce Proximal Policy Optimization Algorithms~\cite{schulman2017proximal} (PPO) with masks to apply the supervision. 
The tuning loss $L_{rl}$ is presented as a reward in policy gradient as defined below:
\begin{equation}
\label{eq:adloss}
	\begin{aligned}
		\cL_{rl}=LLM(\bx^{T_i}|,\bx^{1:T_i-1})R(\bx^{1:T_i},mask),
	\end{aligned}
\end{equation}
where mask denotes whether this is a parameter token. If the token is not a parameter token the reward will be always 0.  $\bx^{T_i}$ denotes the predicted token at time step $T_i$ and $R(\bx^{1:T_i})$ denotes the reward function for waypoint prediction and safety metrics. 
Because there is no sense in letting discriminators only judge
on an incomplete sequence, Monte Carlo search (MC search) is applied to the sample and average unknown $|x|-T_i$ tokens (where $|x|$ is the total length).
\begin{align}
\label{eq:mcreward}
R(x^{1:T_i},mask) = 
\begin{cases}
\frac{1}{N} \sum_{n=1}^N \left\|\bw_t - \bw_t^{gt}\right\|_1   \\
\cL_{safety}(x^{1:T_i}), \quad x^{1:T_i} \in safety
\end{cases}
\end{align}

where $\bx_n^{1:|x|}$ denotes $n^{th}$ sampled sequence by MC search, and $\cL_{safety}$ is the safety metrics used in InterFuser~\cite{shao2022safety}. 
The training loss we use for training is the combination of normal language loss and reward loss, $L=L_{MLE} + \lambda L_{rl}.$

%% file: text/experiment.tex
\section{Experiment}

\subsection{Experimental Setting}\label{exp:setting}
\noindent\textbf{Simulation Environment:} 
This paper performs the end-to-end driving task under the simulator. 
We use the \textbf{CARLA Simulator 0.9.14}~\footnote{\href{https://carla.org/}{https://carla.org/}}, a high-performance driving simulator under urban scenario as our simulation environment. 
The driving is performed in a variety of areas under various weather and lighting conditions. 
The agent is expected to perform driving given predefined routes under complicated traffic and weather conditions, where the routes are a sequence of sparse goal locations with GPS coordinates. 


\noindent\textbf{Data Collection:}
We use the simulator to collect 1000 routes in 8 official town maps with an average length of 400m by using CARLA rule-based expert auto-agent~\cite{shao2023safety} with 228k training pairs.

\noindent\textbf{Evaluation Benchmark:}
This paper is evaluated by both online benchmark and offline benchmarks. 
Although the CARLA simulator provides an official evaluation leaderboard, the use time is restricted to only 200 hours per month yet K80 GPUs are not feasible to deploy large LLMs, which makes the official leaderboard unsuitable for ablation studies or obtaining detailed statistics involving
We conduct our ablation comparison based on the Longeset6 Benchmark proposed by Transfuser~\cite{transfuser}.

\noindent\textbf{Evaluation Metrics:}
For both online evaluation and offline evaluation, we follow the official evaluation metrics to calculate three main metrics, \textit{Route Completion} (RC), \textit{Infraction Score} (IS), and \textit{Driving Score} (DS).
The RC score is the percentage of route distance
completed. Given $R_i$ as the completion by the agent in route $i$, RC is calculated by averaging the completion rate $RC = \frac{1}{N} \sum_i^N R_i$. 
The IS is calculated by accumulating the infractions $P_i$ incurred by the agent during completing the routes. 

\subsection{Implementation Details}\label{exp:implementation}

\textbf{Architectual:} The multi-modality hybrid fusion network uses cameras and LiDAR as the sensory inputs, where image input is monotonic with FOV 120 degree and reshaped into shape $(160, 704)$ and the LiDAR point cloud is converted to BEV representation~\cite{pointpillars} and reshaped into $256, 256$. 
We apply angular viewpoint augmentation~\cite{transfuser} for the training data, by randomly rotating the LiDAR inputs by ±20◦ and adjusting ground truth labels accordingly.
Self-attention modules are applied between feature maps after the first two convolutional blocks, where the resolution is respectively $(C_1, 40, 176)$ and $(C_2, 20, 88)$, where $C_1, C_2$ denotes the channel dimension $72,216$ under our setting. 
For the attention module, two transformer layers with hidden dimensions $512$ and $4$ attention heads are respectively applied as encoder and decoder sequentially.
For polar ray grid sampling details, we follow the settings of previous work~\cite{saha2022translating}. 
The joint token encoding is applied on feature maps with resolutions $20,88$ and $32,32$ after two times downsampling. 
InterFuser~\cite{shao2022safety} and TCP~\cite{wu2022trajectoryguided}.
For controller and safety loss, we directly follow structures from InterFuser~\cite{shao2022safety}. 

Training of the LLMs: The driving data are gathered by autonomous agents as detailed in \cite{shao2022safety}, ensuring comprehensive logging of all safety metrics. These metrics are then transformed into "driving languages" through a set of predefined prompts, which utilize task token 1 for representation. Subsequently, the driving model is trained employing the loss functions delineated in InterFuser~\cite{shao2022safety}. Error data are compiled by assessing the training set, capturing all outputs from both the driving model and the ground truth (GT). Notably, these driving languages incorporate task tokens 2 and 3 at sampling rates of $20\%$ and $80\%$, respectively. The uncertainty prediction layer undergoes training by categorizing this duo of data into binary labels and then leveraging the cross-entropy loss for optimization. As for the language model, we employ Vicuna 33B~\footnote{\href{https://huggingface.co/lmsys/vicuna-33b-v1.3}{https://huggingface.co/lmsys/vicuna-33b-v1.3}} as the pre-trained checkpoints and perform finetuning

\begin{table}[hbpt]
\caption{\label{tb:longset6}Local Driving Evaluation on LongSet6 on Carla 0.9.14}
\centering
\resizebox{0.95\columnwidth}{!}{
\begin{tabular}{lccc}
\Xhline{1.2pt}
  Method               & \textbf{DS $\uparrow$}          & \textbf{RC    $\uparrow$ $\scriptsize{(\%)}$}      & \textbf{IS$\uparrow$}       \\ \hline
NEAT~\cite{chitta2021neat}             &  20.63±{2.34}    & 45.31±{3.25}  &   0.54±{0.03}          \\
WOR~\cite{chen2021learning}              & 21.48±{2.09}  &  52.28±{4.07} &   0.51±{0.05}  \\
GRIAD~\cite{chekroun2021gri}            & 26.98±{4.23}  & 71.43±{3.57}      &    {0.56±{0.04}}   \\
LAV~\cite{chen2022learning}     & 37.62±{1.77}    &  83.94±{2.69}  &   {0.59±{0.05}}        \\ 
GF~\cite{transfuser}  &   27.44±{2.11}  &   80.43±{3.98}   &    0.32±{0.02}         \\
Late TF~\cite{transfuser}  &   35.42±{4.07}  &   83.13±{1.01}   &    0.39±{0.04}         \\
TransFuser~\cite{transfuser}       & {46.95±{5.81}} & {89.64±{2.08}} & 0.52±{0.08} \\

TCP~\cite{wu2022trajectoryguided}       & {49.95±{3.78}} & {92.12±{2.08}} & 0.59±{0.08} \\
InterFuser~\cite{shao2022safety}       & \textbf{52.68±{6.23}} & {92.08±{2.78}} & \textbf{0.62±{0.09}} \\ \hline
PromptGPT        &  52.34±{5.11}   &   \textbf{92.37±{1.09}}  &   {0.60±{0.09}}        \\
\Xhline{0.8pt}
Expert           &  75.83±{2.45}   &   89.82±{0.59}  &  0.85±{0.03}      \\ \Xhline{1.2pt}
\end{tabular}}

\end{table}

\subsection{Driving Performance}\label{exp:drivingscore}

\textbf{Comparison Baseline:} We compare this method with state-of-the-art (SOTA) E2E driving baselines below: 1) NEAT~\cite{chitta2021neat} the attention field fusion on image inputs.
2) WOR~\cite{chen2021learning}, multi-stage Q-function based learning framework; 3) GRIAD~\cite{chekroun2021gri} general reinforced imitation baseline; 4) LAV~\cite{chen2022learning}, the extension of WOR learning from multiple observed agents; 5) Transfuser~\cite{transfuser}, the solid SOTA baseline which introduce transformer fusion into E2E driving. 
Since MMFN~\cite{zhang2022mmfn} are Transfuser baseline if only with image and Lidar, we do not list the evaluation results of MMFN.
Since the target of this paper is to enhance the modality fusion, we also do not list InterFuser~\cite{shao2022safety} as it introduces an additional controller and decision module. 
We also include Late-Transfuser (Late TF) and geometrical fusion (GF) in~\cite{transfuser} to report the early-late fusion comparison. 
TCP~\cite{wu2022trajectoryguided} and InterFuser~\cite{shao2022safety} are selected as the representative of adding additional safety guidance. 

\textbf{Metrics Analysis:} The performance is reported in the offline benchmark LongSet6~\cite{transfuser} in Table~\ref{tb:longset6}. 
The mean and variance metric values are calculated by evaluating each agent three times. 
PromptGPT reached a mean driving score (DS) of $49.08$ and route completion (RC) of $91.87$. 
Although the general performance is lower than current SOTA baselines InterFuser~\cite{shao2022safety} and TCP~\cite{wu2022trajectoryguided}, the PromptGPT-based driving could still reach competitive performance compared to other previous methods. 
Meanwhile, the safety score (IS) is with the same level of TCP while route completion (RC) is lower, which is rational since comprehensive language-based driving could better consider the corner cases.

\begin{figure}
    \centering
    \includegraphics[width=0.95\columnwidth]{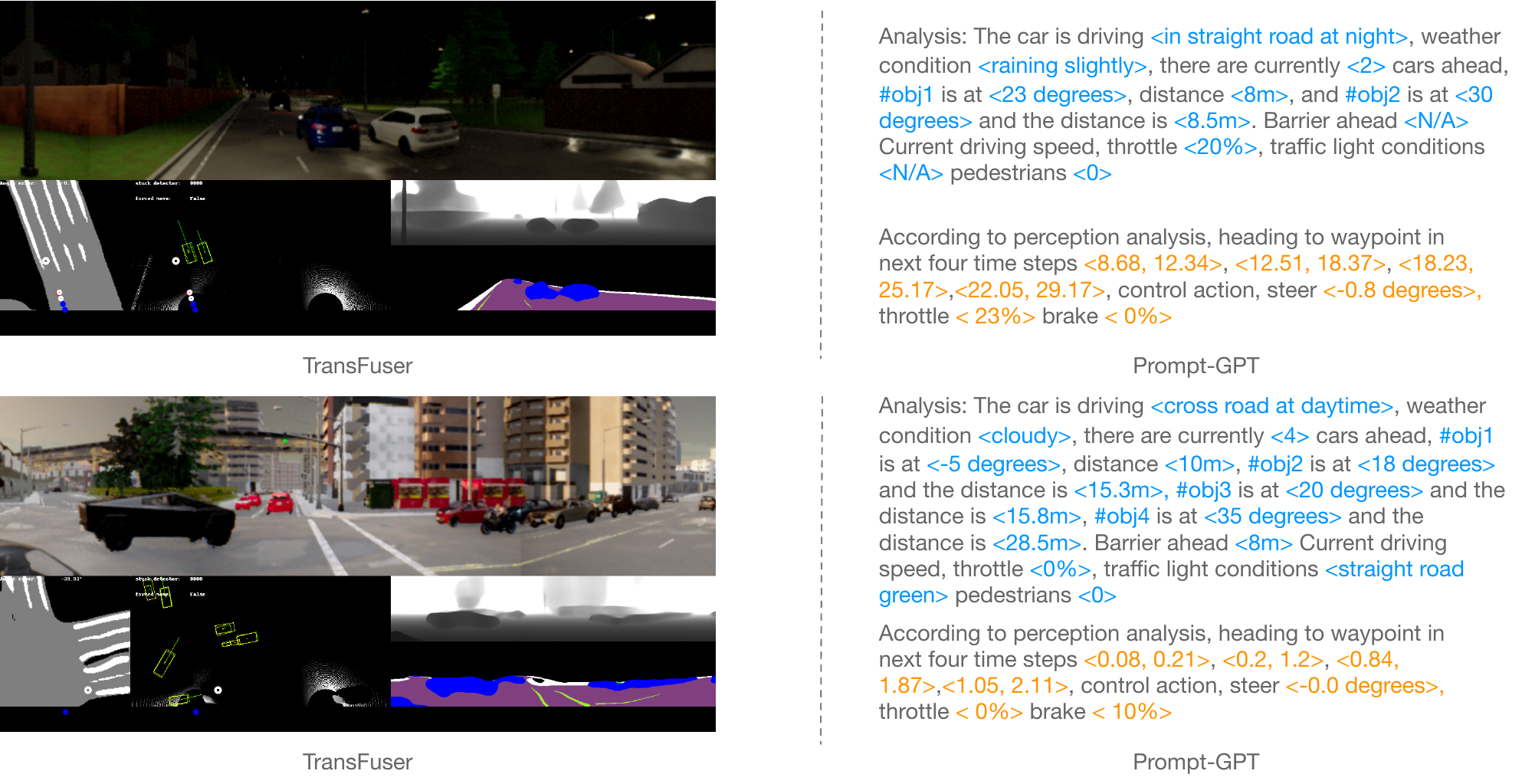}
    \caption{Visualization of driving states between traditional methods and language model. }
    \label{fig:visualization}
\end{figure}

\textbf{Driving Visualization}
To give more intuitive explanations, we visualize the driving mediate state for both traditional waypoints prediction networks (TransFuser~\cite{transfuser} and LLM-based E2E driving methods in Fig.~\ref{fig:visualization}. 
Given two typical scenarios, we could observe that language prompts could describe most of the information in the driving scenario. Yet, it still suggests that LLMs are still not very good at recognizing small and partially visible vehicles. 

\subsection{Ablation Study}

To assess the impact of LLM-facilitated driving correction, we conduct driving experiments across three distinct tasks:
1) Driving solely with LLMs,
2) Driving exclusively with the driving model, and
3) Employing both LLMs and the driving model for driving.
Furthermore, we introduce various scales of LLMs to evaluate the influence of model size on performance. The results are presented in Table~\ref{tb:llm_driving_abl}

\begin{table}[hbpt]
\caption{ \label{tb:llm_driving_abl} Ablation study on different settings for DriveLM.  }
\begin{tabular}{llcc}
\Xhline{1.2pt}
Component & Method & \textbf{DS $\uparrow$} & \textbf{RC    $\uparrow$ $\scriptsize{(\%)}$} \\ \hline
\multirow{3}{*}{LLMs} & Vicuna 7B & 16.54±{2.34} & 39.28±{3.42} \\
 & Vicuna 13B & 34.55±{2.21} & 79.43±{4.05} \\
 & Vicuna 33B & 43.08±{5.28} & 91.87±{1.08} \\
 \hline
Task-Type & Drive (Vicuna 33B) & 43.08±{5.28} & 87.43±{3.98} \\
 & Drive + w/o re-query & 39.62±{4.30} & 83.44±{2.03} \\
 & Drive + Correction & 52.34±{5.11} & 92.37±{1.09} \\ \Xhline{1.2pt}
\end{tabular}
\end{table}

Our observations indicate that when relying solely on language model-based driving, the performance attains a driving score of $43.08$. This is a reduction of $9.26$ compared to driving with a pure language model. When using the initially generated output directly, the performance further declines to $39.62$. However, when employing LLMs as correction advisers, the performance elevates to state-of-the-art levels.
These results underscore the effectiveness of our approach. Additionally, concerning the influence of model scales, there's a distinct correlation between model size and performance. As we scale the LLM from 7B to 33B, there's a notable performance boost. This trend aligns with our expectations, as intricate driving logic necessitates substantial model capacity.

\section{Limitation}
While language models exhibit competitive performance in driving simulators, they are not without limitations:
The inference speed, combined with the re-query mechanism, results in sluggish response times, rendering it unsuitable for real-time driving. Additionally, the complexity of the driving framework poses challenges for evaluation on online benchmarks. 
The unpredictability inherent in language prompts is also a concern. On occasion, the LLM may not generate the necessary driving actions, prematurely halting and solely predicting perception parameters.
Also, in order to train LLMs for perform correction, the current framework need the driving model pre-trained and evaluated in a `off-shelf' formation to collect language training corpus, which is not yet End-to-End as well as heavy to train. 

\section{Conclusion}
\label{sec:conclusion}

In this work, we delve into the innovative approach of utilizing LLMs with multi-modality tokens for end-to-end (E2E) autonomous driving. Rather than letting LLMs take the wheel directly, we suggest harnessing them to aid driving models in rectifying erroneous behaviors, especially in complex scenarios. Our experiments indicate that this corrective approach can prevent LLMs from generating non-existent actions. While the current language model hasn't yet surpassed state-of-the-art performance, it demonstrates a commendable level of competitiveness. This implies the significant potential for integrating LLMs within the realm of autonomous driving.

